\documentclass[11pt,a4paper]{article}
\usepackage[hyperref]{emnlp2020}
\pdfoutput=1
\usepackage{times}
\usepackage{latexsym}

\usepackage{amsmath}
\usepackage{url}
\usepackage{amssymb}
\usepackage{amsfonts}
\usepackage{graphicx}
\usepackage{tabularx}
\usepackage{multirow}
\usepackage{arydshln}
\usepackage{mathtools,nccmath}

\usepackage[utf8]{inputenc}
\usepackage[utf8]{vietnam}
\usepackage{enumitem}

\usepackage{todonotes}

\usepackage{microtype}

\aclfinalcopy 

\title{A Pilot Study of Text-to-SQL Semantic Parsing for Vietnamese}

\author{Anh Tuan Nguyen$^{1,}$\thanks{\ \ Work done during internship at  VinAI Research.} , Mai Hoang Dao$^2$ \and Dat Quoc Nguyen$^2$\\
 $^1$NVIDIA, USA;  $^2$VinAI Research, Vietnam\\
   \tt{\normalsize tuananhn@nvidia.com, \{v.maidh3, v.datnq9\}@vinai.io}}

\date{}

\begin{document}
\maketitle
\begin{abstract}
Semantic parsing is an important NLP task. However, Vietnamese is a low-resource language in this research area. In this paper, we present the \emph{first} public large-scale Text-to-SQL semantic parsing dataset for Vietnamese. We extend and  evaluate two strong semantic parsing baselines EditSQL \citep{ZhangYESXLSXSR19} and  IRNet  \citep{abs-1905-08205} on our dataset. We  compare the two baselines with key configurations and find that: automatic Vietnamese word segmentation  improves the parsing results of both baselines; the normalized pointwise mutual information (NPMI) score \citep{bouma2009normalized} is useful for schema linking;  latent syntactic features extracted from a neural dependency parser for Vietnamese also improve the results; and the monolingual language model PhoBERT for Vietnamese \citep{abs-2003-00744} helps produce higher performances than the recent best multilingual language model XLM-R \citep{abs-1911-02116}.
\end{abstract}

\section{Introduction}

Semantic parsing is the task of converting natural language sentences into  meaning representations such as logical forms or standard SQL database queries  \citep{10.1007/978-3-540-70939-8_28}, which serves as an important component in many NLP systems  such as Question answering and Task-oriented dialogue  \citep{androutsopoulos_ritchie_thanisch_1995,moldovan-etal-2003-cogex,NIPS2018_7558}. 
The significant availability of  the world's knowledge stored in relational databases leads to the creation of large-scale Text-to-SQL datasets, such as WikiSQL \citep{abs-1709-00103} and Spider \citep{YuZYYWLMLYRZR18}, which help boost the development of various state-of-the-art sequence-to-sequence (seq2seq) semantic parsers \citep{abs-1908-11214,ZhangYESXLSXSR19,abs-1905-08205}. Compared to WikiSQL, the Spider dataset presents challenges not only in handling complex questions but also in generalizing to unseen databases during evaluation. 

Most SQL semantic parsing benchmarks, such as WikiSQL and Spider, are exclusively for English. Thus the development of semantic parsers has largely been limited to the English language. As SQL is a database interface and universal semantic representation, it is worth investigating  the Text-to-SQL semantic parsing task for languages other than English. Especially, the difference in linguistic characteristics could add difficulties in applying  seq2seq semantic parsing models to the non-English languages \citep{abs-1909-13293}. 
For example, about 85\% of word types in Vietnamese are composed of at least two syllables \citep{DinhQuangThang2008}. Unlike English, in addition to marking word boundaries, white space is also used to separate syllables that constitute words in Vietnamese written texts. For example, an 8-syllable written text ``Có bao nhiêu quốc gia ở châu Âu'' (How many countries in Europe) forms 5 words ``Có bao\_nhiêu\textsubscript{How many} quốc\_gia\textsubscript{country} ở\textsubscript{in} châu\_Âu\textsubscript{Europe}''. Thus it is interesting to study the influence of word segmentation in Vietnamese on its SQL parsing, i.e. syllable level vs. word level. 

In terms of Vietnamese semantic parsing, previous approaches construct  rule templates to convert single database-driven questions into meaning representations \citep{huongnlidb,5361736,NguyenIEAAIE2012b,TungMH15,NguyenNP_SWJ}. Recently,  \newcite{VuongNTNP19} formulate the Text-to-SQL semantic parsing task for Vietnamese as a sequence labeling-based slot filling problem, and then solve it by using a conventional CRF model with handcrafted features, due to the simple structure of the input questions they deal with.  Note that seq2seq-based semantic parsers have not yet been explored in any previous work w.r.t. Vietnamese. 

Semantic parsing datasets for Vietnamese include a corpus of 5460 sentences for assigning semantic roles \cite{JCSCEvisp} and a small Text-to-SQL dataset of 1258 simple structured questions over 3 databases \cite{VuongNTNP19}. However, these two datasets are not publicly available for research community.

In this paper, we introduce the \emph{first} public large-scale Text-to-SQL dataset for the Vietnamese semantic parsing task. In particular, we create this dataset by manually translating the Spider dataset into Vietnamese. We empirically evaluate strong seq2seq baseline parsers EditSQL \citep{ZhangYESXLSXSR19} and IRNet  \citep{abs-1905-08205} on our dataset. 

Extending the baselines, we extensively investigate key configurations and find that: (1) Our human-translated dataset is far more reliable  than a dataset consisting of machine-translated questions, and the overall  result obtained for Vietnamese is comparable to that for English. (2) Automatic Vietnamese word segmentation improves the performances of the baselines. (3) The NPMI score \citep{bouma2009normalized}  is useful for linking a cell value mentioned in a question to a column in the database schema. 
(4) Latent syntactic features, which are dumped from a neural  dependency parser pre-trained for Vietnamese \citep{nguyen-verspoor-2018},  also help improve the performances. (5) Highest improvements are accounted for the  use of pre-trained language models, where PhoBERT \citep{abs-2003-00744} helps produce higher results than  XLM-R \citep{abs-1911-02116}.

%
%
%
%
%
%
We hope that our dataset can serve as a starting point for future Vietnamese semantic parsing research and applications. 
We publicly release our dataset at: \url{https://github.com/VinAIResearch/ViText2SQL}.

\section{Our Dataset}

We manually translate all English questions and the database schema (i.e. table and column names as well as values in SQL queries)  in  Spider   into Vietnamese. Note that the original Spider dataset consists of 10181 questions with their corresponding 5693 SQL queries over 200 databases. However, only 9691 questions and their corresponding 5263 SQL queries over 166 databases, which are used for training and development, are publicly available. Thus we could only translate those available ones. 

The translation work is performed by 1 NLP researcher and 2 computer science students (IELTS 7.0+). Every question and SQL query pair from the same database is first translated by one student and then cross-checked and corrected by the second student; and finally the NLP researcher verifies the original and corrected versions and makes  further revisions if needed. Note that in case  we have literal translation for a question, we stick to the style of the original English question as much as possible. Otherwise, for complex questions, we will rephrase them based on the semantic meaning of the corresponding SQL queries to obtain the most natural language questions in Vietnamese.

Following \newcite{YuZYYWLMLYRZR18} and \newcite{abs-1909-13293}, we split our dataset into training, development and test sets such that no database overlaps between them, as detailed in Table \ref{tab:data}. Examples of question and SQL query pairs from our dataset are presented in Table \ref{tab:examples}. 
Note that translated question and SQL query pairs in our dataset are written at the syllable level. To obtain a word-level version of the dataset, we apply RDRSegmenter \citep{nguyen-etal-2018-fast} from VnCoreNLP  \citep{vu-etal-2018-vncorenlp} to perform automatic Vietnamese word segmentation.

\begin{table}[!t]
    \centering
    
    \resizebox{7.75cm}{!}{
    \setlength{\tabcolsep}{0.25em}
    \begin{tabular}{l|lllc|llll}
    \hline
    \textbf{}  & \textbf{\#Qu.} & \textbf{\#SQL} & \textbf{\#DB} & \textbf{\#T/D} & \textbf{\#Easy} & \textbf{\#Med.} & \textbf{\#Hard} & \textbf{\#ExH} \\
    \hline
    all & 9691 & 5263 & 166 & 5.3 & 2233 & 3439 & 2095 & 1924 \\
    \hdashline
    train & 6831 & 3493 & 99 & 5.4 & 1559 & 2255 & 1502 & 1515 \\
    dev & 954 & 589 & 25 & 4.2 & 249 & 405 & 191 & 109 \\
    test & 1906 & 1193 & 42 & 5.7 & 425 & 779 & 402 & 300 \\
    \hline
    \end{tabular}
    }
    \caption{Statistics of our human-translated dataset.  ``\#Qu.'', ``\#SQL'' and ``\#DB'' denote  the numbers of questions, SQL queries and databases, respectively. ``\#T/D'' abbreviates the average number of tables per database. ``\#Easy'', ``\#Med.'', ``\#Hard'' and ``\#ExH'' denote the numbers of questions categorized by their SQL queries' hardness levels of ``easy'', ``medium'', ``hard'' and ``extra hard'', respectively (as defined by \citeauthor{YuZYYWLMLYRZR18}).  }
    \label{tab:data}
\end{table}

\begin{table}[!ht]
    \centering
     \resizebox{7.75cm}{!}{
     \begin{tabular}{l}
    \hline
    \textbf{Original} ({Easy question}--involving one table in one database):  \\
    What is the number of cars with more than 4 cylinders? \\
     SELECT count(*) FROM CARS\_DATA WHERE Cylinders  >  4 \\
   \hdashline
    \textbf{Translated}: \\
    Cho biết số lượng những chiếc xe có nhiều hơn 4 xi lanh. \\
    SELECT count(*) FROM [dữ liệu xe] WHERE [số lượng xi lanh]  >  4 \\
    
    \hline
    \hline
    \textbf{Original} ({Hard question}--with a nested SQL query): \\
    Which countries in europe have at least 3 car manufacturers? \\
    SELECT T1.CountryName FROM COUNTRIES AS T1 JOIN CONTINENTS \\
    AS T2 ON T1.Continent  =  T2.ContId JOIN CAR\_MAKERS \\
    AS T3 ON T1.CountryId  =  T3.Country \\
    WHERE T2.Continent  =  ``europe'' GROUP BY T1.CountryName \\
    \ \ \ \ HAVING count(*)  >=  3 \\
    \hdashline
    \textbf{Translated}: \\
    Những quốc gia nào ở châu Âu có ít nhất 3 nhà sản xuất xe hơi? \\
    SELECT T1.[tên quốc gia] FROM [quốc gia] AS T1 JOIN [lục địa] \\ AS T2 ON T1.[lục địa]  =  T2.[id lục địa] JOIN [nhà sản xuất xe hơi] \\AS T3 ON T1.[id quốc gia] = T3.[quốc gia] \\ WHERE T2.[lục địa]  =  ``châu Âu'' GROUP BY T1.[tên quốc gia] \\ 
    \ \ \ \ HAVING count(*)  >=  3 \\
    \hline
     \end{tabular}
    }
    \caption{Syllable-level examples. Word segmentation outputs are not shown for simplification.}
    \label{tab:examples}
 \end{table}


\section{Baseline Models and Extensions}
\subsection{Baselines}

Recent state-of-the-art results on the Spider dataset are reported for RYANSQL \citep{choi2020ryansql} and  RAT-SQL \citep{wang2019ratsql}, 
which are based on the seq2seq encoder-decoder architectures. However, their implementations are not published at the time of our empirical investigation.\footnote{The implementations are still not yet publicly available on 03/06/2020---the EMNLP 2020's submission deadline.}  
Thus we select seq2seq based models EditSQL \citep{ZhangYESXLSXSR19} and  IRNet  \citep{abs-1905-08205} with publicly available implementations as our baselines, which produce near state-of-the-art scores on Spider. 
We briefly describe the baselines EditSQL  and  IRNet  as follows:

\begin{itemize}[leftmargin=*]

\item 
EditSQL is developed for a context-dependent Text-to-SQL parsing task,  consisting of: (1) a BiLSTM-based question-table encoder to explicitly encode the question and table schema, (2) a BiLSTM-based interaction encoder with attention to incorporate the recent question history, and (3) a LSTM-based table-aware decoder with attention, taking into account the outputs of both encoders to generate a SQL query.

\item 
IRNet first performs an n-gram matching-based schema linking to identify the columns and the tables mentioned in a question. Then it takes the question, a database schema and the schema linking results as input to synthesize a tree-structured SemQL query---an intermediate representation bridging the input question and a target SQL query. This synthesizing process is performed by using a BiLSTM-based question encoder and an attention-based schema encoder together with a grammar-based LSTM decoder \citep{yin-neubig-2017-syntactic}. Finally,  IRNet deterministically uses the synthesized SemQL query to infer the SQL query with domain knowledge.

\end{itemize}

See \newcite{ZhangYESXLSXSR19} and  \newcite{abs-1905-08205} for more details of EditSQL and IRNet, respectively.

\subsection{Our Extensions}

\paragraph{NPMI for schema linking:} IRNet essentially relies on the large-scale knowledge graph ConceptNet \citep{SpeerCH17} to link a cell value mentioned in a question to a column in the database schema, based on two ConceptNet categories `is a type of' and `related terms'. However, these two ConceptNet categories are not available for Vietnamese. Thus we  propose a novel use of
the NPMI collocation score \citep{bouma2009normalized}  for the schema linking in IRNet, which ranks the NPMI scores between the cell values and column names to match a cell value to its column. 

\paragraph{Latent syntactic features:} Previous works have shown that syntactic features  help improve semantic parsing \citep{Monroe2014DependencyPF,jie-lu-2018-dependency}. Unlike these works that use handcrafted syntactic features extracted from dependency parse trees, and inspired by \newcite{zhang-etal-2017-end}'s relation extraction work, we investigate whether latent syntactic features, extracted from the BiLSTM-based dependency parser \texttt{jPTDP} \citep{nguyen-verspoor-2018} pre-trained for Vietnamese, would help improve Vietnamese Text-to-SQL parsing. In particular, our approach is that we dump latent feature representations from \texttt{jPTDP}'s BiLSTM encoder given our word-level inputs, and directly use them as part of input embeddings of EditSQL and IRNet. 

\paragraph{Pre-trained language models:} 
\newcite{ZhangYESXLSXSR19} and  \newcite{abs-1905-08205} make use of BERT \citep{DevlinCLT19} to improve their model performances. Thus we also extend EditSQL and IRNet with the use of pre-trained language  models  XLM-R-base \citep{abs-1911-02116}  and PhoBERT-base \citep{abs-2003-00744} for  the syllable- and word-level settings, respectively. 
XLM-R  is  the recent best multi-lingual model, based on  RoBERTa \citep{RoBERTa}, pre-trained on a 2.5TB multilingual corpus which contains 137GB of syllable-level Vietnamese texts. PhoBERT is a monolingual variant of RoBERTa for Vietnamese, pre-trained on a 20GB of word-level Vietnamese texts.

\setcounter{table}{3}
  \begin{table*}[!t]
    \centering
     \resizebox{16cm}{!}{
     \setlength{\tabcolsep}{0.25em}
     \begin{tabular}{l|cccc|ccccc}
    \hline
    \textbf{Approach} & \textbf{Easy} & \textbf{Medium} & \textbf{Hard} & \textbf{ExH} & \textbf{SELECT} & \textbf{WHERE} & \textbf{ORDER BY} & \textbf{GROUP BY} & \textbf{KEYWORDS} \\
    \hline
    \textbf{EditSQL}\textsubscript{DeP} & 65.7 & 46.1 & 37.6 & 16.8 & 75.1 & 44.6 & 65.6 & 63.2 & 73.5\\
    \textbf{EditSQL}\textsubscript{XLM-R} & 75.1 & 56.2 & 45.3 & 22.4 & 82.7 & 60.3 & 70.7 & 67.2 & 79.8 \\
    \textbf{EditSQL}\textsubscript{PhoBERT} & 75.6 & 58.0 & 47.4 & 22.7 & 83.3 & 61.8 & 72.5 & 67.9 & 80.6 \\
    \hline 
    \textbf{IRNet}\textsubscript{DeP} & 71.8 & 51.5 & 47.4 & 18.5 & 79.3 & 48.7 & 71.8 & 63.4 & 74.3\\
    \textbf{IRNet}\textsubscript{XLM-R} & 76.2 & 57.8 & 46.8 & 23.5 & 83.5 & 59.1 & 74.4 & 68.3 & 80.5\\
    \textbf{IRNet}\textsubscript{PhoBERT} & 76.8 & 57.5 & 47.2 & 24.8 & 84.5 & 59.3 & 76.6 & 68.2 & 80.3\\
    \hline
     \end{tabular}
    }
    \caption{Exact matching accuracy categorized by 4 different hardness levels, and F\textsubscript{1} scores of different SQL components on the test set. ``ExH'' abbreviates Extra Hard.}
    \label{tab:components}
 \end{table*}

\setcounter{table}{2}
\begin{table}[!t]
    \centering
    \resizebox{7.75cm}{!}{
     \setlength{\tabcolsep}{0.3em}
     \def\arraystretch{1.1}
     \begin{tabular}{llll|lll}
    \hline
    & \textbf{Approach} & \textbf{dev} & \textbf{test}& \textbf{Approach} & \textbf{dev} & \textbf{test}\\
    \hline
   \multirow{3}{*}{\rotatebox[origin=c]{90}{\footnotesize{Vi-Syllable}}} 
   & \textbf{EditSQL} [MT]  & 21.5 & 16.8 
   &  \textbf{IRNet} [MT]  & 25.4 & 20.3 \\
   &  \textbf{EditSQL} & 28.6 & 24.1 
   & \textbf{IRNet}  & 43.3 & 38.2 \\
    & \textbf{EditSQL}\textsubscript{XLM-R} & 55.2 & 51.3
    &  \textbf{IRNet}\textsubscript{XLM-R} & 58.6 & 52.8 \\
    \hdashline
    \multirow{4}{*}{\rotatebox[origin=c]{90}{\footnotesize{Vi-Word}}} 
    & \textbf{EditSQL} [MT]  & 22.8 & 17.4 
    & \textbf{IRNet} [MT]  & 27.4 & 21.6 \\
    & \textbf{EditSQL} & 33.7 & 30.2 
    & \textbf{IRNet} & 49.7 & 43.6 \\
    & \textbf{EditSQL}\textsubscript{DeP}  & 45.3 & 42.2 
    & \textbf{IRNet}\textsubscript{DeP}  & 52.2 & 47.1  \\
    & \textbf{EditSQL}\textsubscript{PhoBERT}  & 56.7 & 52.6 
    & \textbf{IRNet}\textsubscript{PhoBERT}  & 60.2 & 53.2 \\
    \hline 
    \hline
    En & \textbf{EditSQL}\textsubscript{RoBERTa} & 58.3 & 53.6 
    & \textbf{IRNet}\textsubscript{RoBERTa} & 63.8 & 55.3 \\
    \hline
    
     \end{tabular}
   }
    \caption{Exact matching accuracies of EditSQL and IRNet. ``Vi-Syllable'' and ``Vi-Word'' denote the results w.r.t. the syllable level and the word level, respectively. \textsc{[MT]}   denotes accuracy results with the machine-translated questions. The subscript ``DeP'' refers to the use of the latent syntactic features. Other subscripts denote the use of the pre-trained language models. ``En'' denotes our results on the  English Spider dataset but under our training/development/test split w.r.t. the total 9691 public available questions.}
    \label{tab:results}
 \end{table}

\section{Experiments}

\subsection{Experimental Setup}

We conduct experiments to study a quantitative comparison between our human-translated dataset and a machine-translated dataset,\footnote{We employ a well-known machine translation engine to translate the English questions into Vietnamese.}  the influence of Vietnamese word segmentation (i.e. syllable level and word level), and the usefulness of the latent syntactic features, the pre-trained language models and the NPMI-based approach for schema linking.  

For both baselines EditSQL and IRNet which require input pre-trained  embeddings for syllables and words, we pre-train a set of 300-dimensional syllable embeddings and another set of 300-dimensional word embeddings using the  Word2Vec skip gram model \citep{MikolovSCCD13} on syllable- and word-level corpora of 20GB Vietnamese texts \citep{abs-2003-00744}.  In addition, we also use these 20GB  syllable- and word-level Vietnamese corpora as our external datasets to compute the NPMI score (with a window size of 20) for schema linking in IRNet.

Our hyperparameters for EditSQL and IRNet are taken from \newcite{ZhangYESXLSXSR19} and  \newcite{abs-1905-08205}, respectively.  The pre-trained syllable and word embeddings are fixed, while the pre-trained language models XLM-R and PhoBERT are fine-tuned during training. 

Following \newcite{YuZYYWLMLYRZR18}, we use two commonly used metrics for evaluation. The first one is the exact matching accuracy, which reports the percentage of input questions that have exactly the same SQL output as its gold reference. The second one is the component matching F\textsubscript{1}, which reports F\textsubscript{1} scores for SELECT, WHERE, ORDER BY,  GROUP BY and all other keywords. 

We run for 10 training epochs and  evaluate the exact matching accuracy after each epoch on the development set, and then select the best model checkpoint to report the final result on the test set.  

\subsection{Main Results}

Table \ref{tab:results} shows the overall exact matching results of EditSQL and IRNet on the development and test sets. Clearly, IRNet does better than EditSQL, which is consistent with results obtained on the original English Spider dataset.  

We find that  our human-translated dataset is far more reliable  than a dataset consisting of machine-translated questions. In particular,  at the word level, compared to the machine-translated dataset, our dataset obtains about 30.2-17.4 $\approx$ 13\% and 43.6-21.6 = 22\% absolute improvements in accuracies of EditSQL and IRNet, respectively (i.e. 75\%--100\% relative  improvements). In addition, the word-based Text-to-SQL parsing obtains about 5+\% absolute higher accuracies than the syllable-based Text-to-SQL parsing (EditSQL: 24.1\%$\rightarrow$30.2\% ; IRNet:  38.2\%$\rightarrow$43.6\%), i.e.  automatic Vietnamese word segmentation improves the accuracy results. 

Furthermore, latent syntactic features dumped from the pre-trained  dependency parser  \texttt{jPTDP} for Vietnamese help improve the performances of the baselines (EditSQL: 30.2\%$\rightarrow$42.2\%; IRNet: 43.6\%$\rightarrow$47.1\%). Also, biggest improvements are accounted for the use of pre-trained language models. In particular, PhoBERT helps produce higher results than  XLM-R (EditSQL: 52.6\% vs. 51.3\%; IRNet: 53.2\% vs. 52.8\%).

We also retrain EditSQL and IRNet on the English Spider dataset with the use of the strong pre-trained language model RoBERTa instead of BERT, but under our dataset split. We find that the overall  results for Vietnamese are smaller but comparable to the English results. Therefore,  Text-to-SQL semantic parsing for Vietnamese might not be significantly more challenging than that for English.

Table \ref{tab:components} shows the exact matching accuracies of EditSQL and IRNet w.r.t. different hardness levels of SQL queries and the F\textsubscript{1} scores w.r.t. different SQL components on the test set. Clearly, in most cases, the pre-trained language models PhoBERT and XLM-R help produce substantially higher results than the latent syntactic features, especially for the \texttt{WHERE} component.

\paragraph{NPMI-based schema linking:} We also investigate the contribution of our NPMI-based extension approach for schema linking in applying IRNet for Vietnamese. Without using NPMI  for schema linking,\footnote{Without schema linking, IRNet assigns a `NONE' type for column names.} we observe 6+\% absolute decrease in the  exact matching accuracies of IRNet on both development and test sets, thus showing the usefulness of our NPMI-based approach for schema linking. 

\subsection{Error Analysis}
To understand the source of errors, we perform an error analysis on the development set which consists of 954 questions. Using   {IRNet}\textsubscript{PhoBERT}  which produces the best result, we identify several causes of errors from 382/954 failed examples.

For 121/382 cases (32\%), IRNet\textsubscript{PhoBERT} makes incorrect predictions on the column names which are not mentioned or only partially mentioned in the questions. For example, given the question ``Hiển thị tên và năm phát hành của những bài hát thuộc về ca sĩ trẻ tuổi nhất'' (Show the name and the release year of the song by the youngest singer),\footnote{Word segmentation is not shown for simplification.} the model produces an incorrect column name prediction of ``tên'' ({name})  instead of the correct one ``tên bài hát'' ({song name}). Errors related to column name predictions can either be missing the entire column names or inserting random column names into the \texttt{WHERE} component of the predicted SQL queries.

About 12\% of  failed examples (47/382) in fact have an equivalent implementation of their intent with a different SQL syntax. For example, the model produces a `failed' SQL output ``\texttt{SELECT MAX} [sức chứa] \texttt{FROM} [sân vận động]'' which is equivalent to the gold SQL query of  ``\texttt{SELECT} [sức chứa] \texttt{FROM} [sân vận động] \texttt{ORDER BY} [sức chứa] \texttt{DESC LIMIT 1}'', i.e. the SQL output would be valid if we measure an execution accuracy.

About 22\% of failed examples (84/382) are caused by  nested and complex SQL queries which mostly belong to the Extra Hard category. With 18\% of failed examples (70/382), incorrectly  predicting operators is another common  type of errors.  For example, given the phrases ``\emph{già nhất}'' (oldest) and ``\emph{trẻ nhất}'' (youngest) in the question, the model fails to predict the correct operators \texttt{max} and \texttt{min}, respectively. The remaining 60/382 cases (16\%) are accounted for an incorrect  prediction of table names in a \texttt{FROM} clause.

\section{Conclusion}

In this paper, we have presented the first public large-scale dataset for Vietnamese Text-to-SQL semantic parsing. We also extensively experiment with key research configurations using two strong baseline models on our dataset and find that: the input representations, the NPMI-based approach for schema linking, the latent syntactic features and the pre-trained language models  all have the influence on this Vietnamese-specific task. 
We hope that our dataset can serve as the  starting point for further research and applications in Vietnamese question answering and dialogue systems.

\bibliography{emnlp2020}
\bibliographystyle{acl_natbib}

\end{document}